\documentclass[sigconf]{acmart}
\usepackage{multirow}
\usepackage{graphicx}
\usepackage{subcaption}
\usepackage{booktabs}
\usepackage{balance}
\AtBeginDocument{%
  }

\begin{document}

\title{SDO: Subspace Deconflicting Operator for Multi-Adapter Composition}

\author{Zhongsheng Wang}
\email{zhongsheng.wang@auckland.ac.nz}
\orcid{0009-0003-4235-7710}
\affiliation{%
  \institution{University of Auckland}
  \city{Auckland}
  \country{New Zealand}
}

\author{Zhedong Lin}
\orcid{0009-0003-5079-9850}
\affiliation{%
  \institution{University of Auckland}
  \city{Auckland}
  \country{New Zealand}
}
\email{zlin629@aucklanduni.ac.nz}

\author{Qian Liu}
\orcid{0000-0002-3162-935X}
\affiliation{%
  \institution{University of Auckland}
  \city{Auckland}
  \country{New Zealand}
}
\email{liu.qian@auckland.ac.nz}

\author{Xinyu Zhang}
\orcid{0000-0002-2999-3291}
\affiliation{%
  \institution{University of Auckland}
  \city{Auckland}
  \country{New Zealand}
}
\email{xinyu.zhang@auckland.ac.nz}

\author{Jiamou Liu}
\orcid{0000-0002-0824-0899}
\correspondingauthor
\affiliation{%
  \institution{University of Auckland}
  \city{Auckland}
  \country{New Zealand}
}
\email{jiamou.liu@auckland.ac.nz}

\renewcommand{\shortauthors}{Zhongsheng Wang, Zhedong Lin, Qian Liu, Xinyu Zhang, and Jiamou Liu}

\begin{abstract}
Composing independently trained adapters within a shared diffusion backbone provides a modular approach to multi-character generation, but naive joint deployment often causes identity mixing, cross-character attribute leakage, and unstable scene composition. We study this interference from a parameter-space perspective and hypothesize that it arises partly from conflicts between overlapping dominant subspaces in shared layers. To address this issue, we propose \textbf{SDO}, a \textbf{S}ubspace \textbf{D}econflicting \textbf{O}perator for multi-adapter composition. SDO reconstructs layer-wise low-rank updates from the selected adapters, extracts compact subspace signatures, measures pairwise conflict through output-subspace overlap, and applies a permutation-equivariant transformation that suppresses harmful shared directions while retaining identity-specific characteristics. The resulting representations are mapped back to standard adapter updates and can be directly incorporated into existing diffusion inference pipelines. Experiments demonstrate that SDO consistently improves identity fidelity and compositional stability, with particularly clear gains as the number of jointly composed adapters increases.

\end{abstract}

\begin{teaserfigure}
  \centering
  \includegraphics[width=0.8\textwidth]{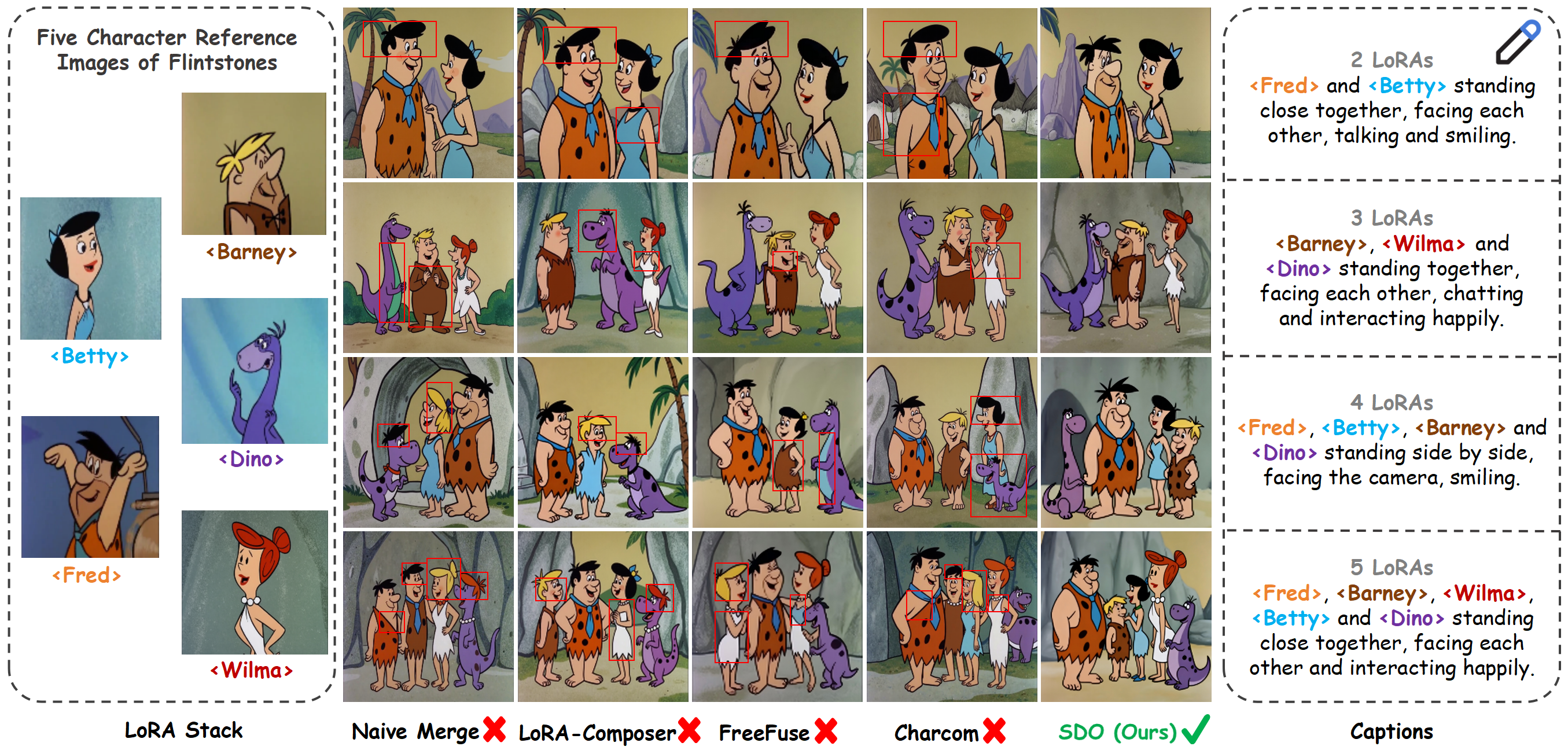}
  \caption{Qualitative comparison on Flintstones multi-character generation with 2 to 5 jointly activated adapters. Each row corresponds to an increasing number of characters, and each column compares different composition methods. Red boxes highlight typical failure cases such as identity mixing and attribute leakage. As scene complexity increases, naive merging and prior methods progressively suffer from degraded identity fidelity and cross-character interference. In contrast, SDO consistently preserves character identity, improves separation, and maintains coherent multi-character interactions.}
  \label{fig:teaser}
\end{teaserfigure}


\maketitle

\section{Introduction}

Multi-character generation is a central yet fragile setting in personalized text-to-image synthesis. In applications such as story illustration, character-centered visual design, and personalized content creation, a model must jointly render multiple identities within a coherent scene while preserving identity consistency, visual separation, and stable representation as scene complexity increases~\cite{lin2026narratology,avrahami2024chosen,wang2024ms}. In this setting, the primary challenge is no longer single-subject realism, but \emph{compositional identity consistency}: each character should remain recognizable, attribute leakage should be minimized, and the scene should remain stable under joint generation.

A common approach is to maintain a pool of independently trained, lightweight adapters, each encoding a specific character, and to compose a selected subset at inference time. This modular paradigm is attractive due to its parameter efficiency, reusability, and flexibility in supporting plug-and-play composition~\cite{hu2021lora,gu2023mixofshow,zhong2024multilora}. However, directly composing independently trained adapters within a shared diffusion backbone is often unreliable. Prior work has reported failure modes such as identity mixing, cross-character attribute leakage, missing subjects, and unstable composition~\cite{gu2023mixofshow,li2024loracomposer}. As illustrated in Fig.~\ref{fig:teaser}, these issues become increasingly severe as more characters are jointly generated.

Existing methods address these failures primarily through conditioning strategies, attention control, spatial guidance, or inference-time heuristics~\cite{gu2023mixofshow,liu2023cones2,xiao2025fastcomposer,wang2024msdiffusion,liu2025freefuse,zhong2024multilora,yang2025loracomposer,meral2024clora,wang2025charcom}. While effective in specific regimes, these approaches mainly regulate interactions at the feature or decoding level. They do not directly address a more fundamental issue: when multiple independently trained adapters are injected into a shared backbone, their parameter updates may already be incompatible.

This observation raises a central question for identity-consistent multi-character generation: \emph{how can independently trained adapters be composed such that identity interference is minimized under shared-backbone generation?} We approach this problem from a parameter-space perspective. 
Since each adapter contributes a structured update to the backbone parameters, their joint deployment can be viewed as composing multiple weight updates within shared layers. 
From this perspective, adapter interference is not merely a prompting or attention-allocation issue, but also a \emph{subspace conflict} problem. 
Incompatible adapters may exhibit overlapping dominant directions in shared layers, leading to competition for similar representational capacity during joint inference. This view connects multi-character generation to prior work on adapter fusion, model merging~\cite{pfeiffer2021adapterfusion,ilharco2023task,yadav2023ties}, and subspace separation~\cite{bousmalis2016dsn,misra2016crossstitch}, where compatibility in parameter space is critical for reliable composition.

Based on this perspective, we propose \textbf{SDO}, a \textbf{S}ubspace \textbf{D}econflicting \textbf{O}perator for multi-adapter composition. Rather than summing adapter weights, SDO reconstructs layer-wise updates, extracts compact subspace signatures, estimates pairwise conflict, and applies a permutation-equivariant transformation to suppress harmful overlap while preserving identity-specific behavior. The transformed representations are then mapped back to standard adapter form, enabling integration with existing diffusion pipelines.

Experiments on multi-character personalized generation show that SDO improves identity fidelity, character separation, and compositional stability. The gains are most pronounced in challenging 4- and 5-character settings, where naive composition and prior methods suffer from severe identity mixing and attribute leakage. These results suggest that explicit subspace-level deconfliction provides a practical route toward identity-consistent multi-adapter composition. Our main contributions are as follows:
\begin{itemize}
    \item We formulate interference in multi-adapter composition as a \emph{subspace conflict} problem, providing a parameter-space perspective on identity mixing and attribute leakage.
    \item We propose \textbf{SDO}, a permutation-equivariant operator that rewrites a selected adapter set in a low-rank signature space and maps the result back to deployable adapter updates.
    \item Extensive experiments demonstrate that SDO improves identity fidelity, separation, and compositional stability, with the largest gains in challenging multi-adapter regimes.
\end{itemize}

\section{Related Work}

\subsection{Personalization and Multi-Adapter Composition}

Personalized diffusion has progressed from modeling a single identity to composing multiple learned identities within one scene. Early methods such as DreamBooth~\cite{ruiz2023dreambooth}, Textual Inversion~\cite{gal2023textualinversion}, and Custom Diffusion~\cite{kumari2023customdiffusion} established subject-specific personalization through fine-tuning or token learning. Later approaches including Perfusion~\cite{tewel2023perfusion}, BLIP-Diffusion~\cite{li2023blipdiffusion}, IP-Adapter~\cite{ye2023ipadapter}, and PhotoMaker~\cite{li2024photomaker} improved scalability, controllability, and data efficiency, often through image-conditioned or parameter-efficient designs. In parallel, SVDiff~\cite{han2023svdiff} highlighted that diffusion adaptation can often be expressed in a compact low-rank form, suggesting that the geometry of adaptation updates is itself important rather than merely an implementation detail.

These advances mainly address single-subject modeling, leaving open the harder problem of jointly rendering multiple personalized subjects in a coherent scene. To improve multi-subject composition, prior work has explored disentangled subject representations~\cite{liu2023cones2,gu2023mixofshow}, attention and layout control~\cite{xiao2025fastcomposer,helbling2023objectcomposer}, and structured or spatial guidance during denoising~\cite{wang2024msdiffusion,liu2025freefuse, liu2026look, 10.1145/3746027.3758165}. Collectively, these methods substantially improve compositional controllability and reduce some common failures in multi-subject generation. However, they mostly regulate interactions at the conditioning, attention, or decoding level, and therefore do not directly address whether the underlying personalized modules are themselves compatible when jointly loaded into a shared backbone.

More recent work moves closer to our setting by studying the direct composition of multiple personalized LoRAs at inference time~\cite{zhong2024multilora,yang2025loracomposer,meral2024clora,zou2025cmlora,wang2025charcom}. Representative examples include LoRA-Composer~\cite{yang2025loracomposer} and CharCom~\cite{wang2025charcom}, which improve modular personalization through inference-time coordination strategies. This line of work is particularly relevant because it explicitly targets the practical setting where a user maintains a pool of independently trained identity adapters and composes a selected subset on demand. Nevertheless, these methods still mainly regulate how adapters interact during generation, rather than directly rewriting the adapter updates themselves. By contrast, our focus is on incompatibility among the participating adapter updates in parameter space.

\subsection{Subspace Conflict in Adapter Composition}

A large body of work on adapter fusion and model merging suggests that composition quality depends strongly on geometry in weight space. AdapterFusion~\cite{pfeiffer2021adapterfusion} shows that multiple adapters can be combined productively, while Task Arithmetic~\cite{ilharco2023task} treats learned parameter deltas as composable directions. TIES-Merging~\cite{yadav2023ties} further shows that naive combination can fail because overlapping or conflicting updates interfere destructively. For our task, these works provide an important starting point: independently useful parameter updates are not necessarily jointly compatible, and successful composition often requires explicitly reasoning about how update directions interact.

Another related line of work studies interference through subspace separation and orthogonality. Domain Separation Networks~\cite{bousmalis2016dsn} and Cross-Stitch Networks~\cite{misra2016crossstitch} show that separating shared and private directions can improve multi-task representation learning. More recently, orthogonality-based LoRA composition methods such as OSRM~\cite{zhang2025osrm} suggest that reducing overlap among low-rank directions can improve mergeability. Although these methods are not designed for personalized diffusion, they sharpen a key intuition behind our method: harmful interaction between modules can often be understood as overlap among dominant subspaces rather than only as a failure of prompting or attention allocation.

This issue is particularly relevant to multi-character personalized diffusion. Although each adapter is trained independently, multiple adapters are integrated into the same frozen diffusion backbone and must jointly modify shared layers during generation. Existing multi-subject and multi-adapter methods evaluate generation quality, controllability, and identity preservation, but rarely characterize whether the composed adapters occupy conflicting directions in parameter space. SDO addresses this overlooked source of interference. Rather than coordinating adapter interactions solely through prompts, attention mechanisms, or denoising procedures, SDO formulates multi-adapter composition as a subspace-conflict problem. It extracts compact low-rank signatures from the selected adapters, measures pairwise subspace overlap, and transforms their updates to suppress harmful shared components while retaining identity-specific behavior. SDO therefore introduces a parameter-space approach to multi-adapter personalized diffusion that complements prior methods centered on generation-time coordination.

\section{Problem Definition}
\label{sec:problem_definition}
\subsection{Preliminaries}

Let $F_{\theta}$ denote a pretrained text-to-image diffusion model whose backbone parameters remain frozen during inference, and let $\mathcal{L}$ denote the set of layers where adapter updates are injected. Given a pool of independently trained character adapters $\mathcal{H} = \{\Delta_1, \dots, \Delta_n\}$, where each adapter $\Delta_i$ is represented as a collection of layer-wise parameter updates $\Delta_i = \{\Delta W_i^{(\ell)}\}_{\ell \in \mathcal{L}}$, with $\Delta W_i^{(\ell)} \in \mathbb{R}^{d_\ell \times k_\ell}$. Here, $d_\ell$ and $k_\ell$ denote the output and input dimensions of layer $\ell$, respectively. Each adapter encodes a distinct identity-specific concept learned independently.

In multi-character generation, we select a subset of adapter indices $S \subseteq \{1, \dots, n\}$ with $2 \le |S| = m \le n$, and jointly activate the corresponding adapters $\{\Delta_i\}_{i \in S}$ during inference to render multiple character identities in a shared scene. The singleton case $|S|=1$ reduces to standard single-adapter generation and requires no composition. A common baseline composes the selected adapters through a layer-wise weighted summation: $\Delta W_{\text{naive}}^{(\ell)} = \sum_{i \in S} \alpha_i \Delta W_i^{(\ell)}$, where $\alpha_i$ is the loading scale of the adapter $i$ during inference.

\subsection{Conceptual Motivation}

\begin{figure}[htbp]
    \centering
    \includegraphics[width=\linewidth]{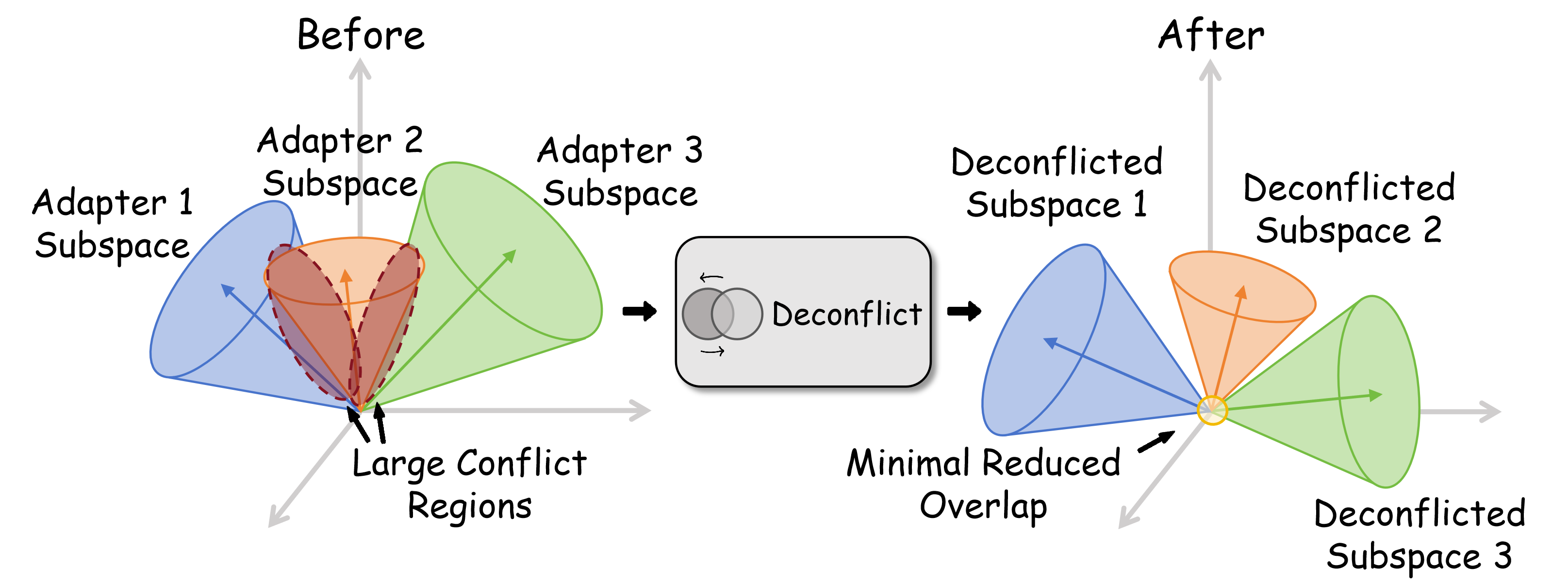}
    \caption{Conceptual illustration of subspace conflict in multi-adapter composition. Before: overlapping dominant directions. After: reduced overlap with preserved identity-specific directions.}
    \label{fig:lora_conflict_vis}
\end{figure}

Naive adapter composition often leads to identity interference during multi-character generation, resulting in identity mixing, cross-character attribute leakage, and unstable joint rendering. These issues become more pronounced as more adapters are jointly activated.

A key contributing factor is that independently trained adapters are optimized in isolation but deployed together within a shared backbone, leading to unintended interactions among their parameter updates during joint inference.

One intuitive way to interpret this mismatch is through a geometric view of the parameter space. Each adapter induces parameter updates that emphasize certain dominant directions in shared layers. When these directions are sufficiently distinct, the adapters tend to remain compatible. In contrast, when multiple adapters exhibit substantial overlap in their dominant directions, their updates may compete, leading to increased interference (see Fig.~2 for a conceptual illustration).

From this perspective, a desirable outcome is to transform the selected adapters into a configuration that reduces such overlap while preserving their identity-specific behaviors. This observation suggests that multi-adapter interference may be related to geometric overlap in parameter space. We therefore hypothesize that treating adapter interaction as a subspace conflict problem provides a useful perspective for designing more stable composition mechanisms.

\begin{figure*}[htbp]
    \centering
    \includegraphics[width=\textwidth]{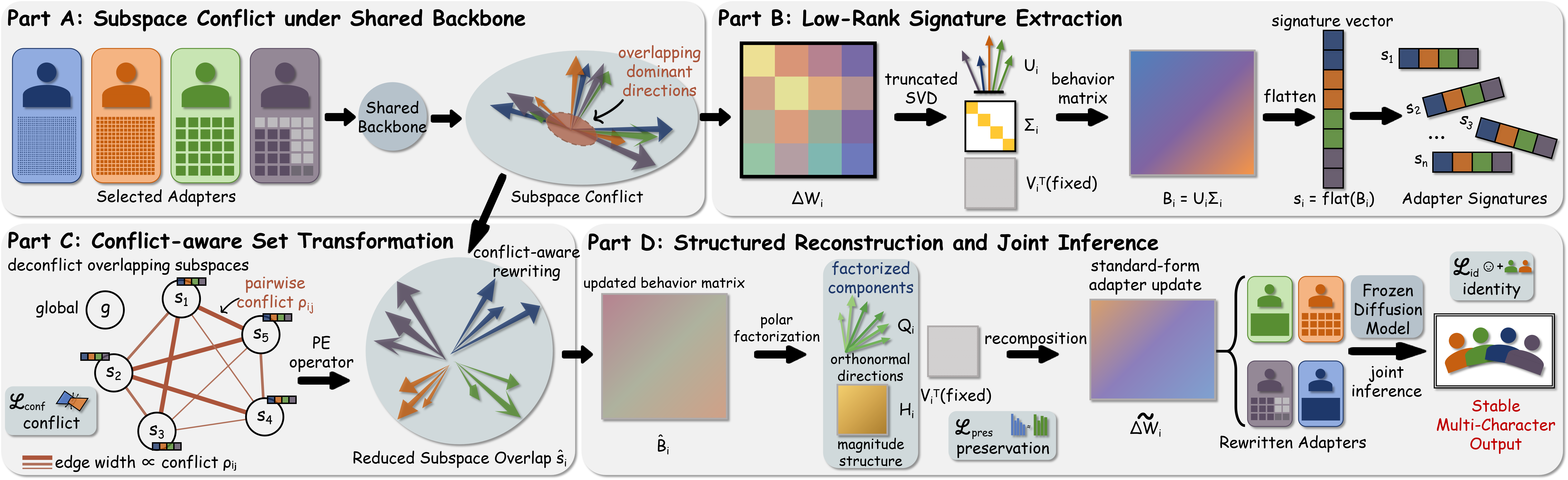}
    \caption{Workflow overview for multi-adapter composition. (A) Independently trained adapters exhibit overlapping dominant directions when injected into a shared backbone, leading to subspace conflict. (B) Each adapter is mapped to a low-rank signature via truncated SVD. (C) A conflict-aware, permutation-equivariant set operator rewrites the signatures to reduce pairwise overlap. (D) The updated representations are reconstructed into standard adapter updates and jointly applied for stable multi-character generation.}
    \label{fig:wide}
\end{figure*}

\subsection{Problem Formulation}

Motivated by these observations, we formulate multi-adapter composition in personalized diffusion as a subspace-conflict problem in parameter space. Given a selected adapter set $\{\Delta_i\}_{i \in S}$, our goal is to learn a set-conditioned transformation operator
$\mathcal{T}: \{\Delta_i\}_{i \in S} \rightarrow \{\widetilde{\Delta}_i\}_{i \in S}$
that rewrites the selected adapters into a new set that is more compatible for joint deployment. Each transformed adapter remains layer-wise, i.e.,
$\widetilde{\Delta}_i = \{\widetilde{\Delta W}_i^{(\ell)}\}_{\ell \in \mathcal{L}}$.

The transformation is required to satisfy the following desiderata:
(1) \textbf{Identity Preservation}: each transformed adapter should retain the identity-specific behavior encoded by its original counterpart.
(2) \textbf{Conflict Mitigation}: the transformed set should exhibit reduced interference under joint composition compared to the original set.
(3) \textbf{Permutation Equivariance}: the transformation should be independent of the ordering of the input adapters, such that reordering the inputs leads to a corresponding reordering of the outputs.

\section{Subspace Deconflicting Operator}
\label{sec:sdo}

\subsection{From Subspace Conflict to SDO}

The key premise behind SDO is that failures in multi-adapter composition are not purely inference-time or prompting artifacts but also arise from geometric conflict in parameter space. When independently trained adapters are injected into the same frozen backbone, their parameter updates act on shared layers rather than as isolated modules. Interference therefore emerges when multiple adapters rely on overlapping dominant directions in those layers.

Under this view, the quantity that should be suppressed is not merely the magnitude of the updates, but the geometric overlap among the dominant adapter-specific directions. We therefore introduce a lightweight pairwise conflict prior on the dominant output-side bases; the formal definition is given later in Eq.~\ref{eq:conflict_score}. Intuitively, larger conflict scores indicate stronger geometric overlap between adapters. Magnitude information is retained in the signature representation and further enforced during optimization through the structure-level objectives.

Because conflict is defined over a selected set of adapters rather than an individual one, resolving it requires a transformation that operates on the set as a whole. This leads to SDO, which rewrites the selected adapters before joint deployment instead of merging them into a single module. At each layer $\ell$, SDO applies a permutation-equivariant, cardinality-preserving transformation to the selected set $X^{(\ell)}$. The transformed representation of each adapter depends on its own state, the surrounding set, and the pairwise conflict relations within that set. This design ensures that each adapter is updated in a context-aware manner, rather than independently.

\subsection{Design Principle of the Operator}

The operator is designed to be set-conditioned and explicitly conflict-aware. Since the input is a set of selected adapters, each transformed output must depend on both its own state and the states of the other adapters. Moreover, as the primary source of interference arises from pairwise geometric incompatibility, the transformation should explicitly incorporate inter-adapter conflict rather than relying solely on learned feature similarity. Finally, because the selected adapters have no canonical order and must remain distinct after transformation, the operator must be permutation-equivariant and cardinality-preserving.

At layer $\ell$, this transformation is instantiated by a layer-wise operator $T_{\psi}^{(\ell)}$ acting on the selected set. Formally, it can be written as
\begin{equation}
  \hat{x}_i^{(\ell)}
  =
  T_{\psi}^{(\ell)}
  \bigl(
    x_i^{(\ell)},
    \{x_j^{(\ell)}\}_{j \neq i},
    \{\rho_{ij}^{(\ell)}\}_{j \neq i}
  \bigr)
\end{equation}
which defines a permutation-equivariant set-to-set mapping where each output is conditioned on its own state, the surrounding set, and the pairwise conflict relations.

This design integrates geometric conflict modeling with set-level transformation, enabling context-aware rewriting of adapters prior to joint deployment.

\section{Implementation Workflow}
\label{sec:method}

\subsection{Workflow Overview}

Given a selected adapter set $S$, SDO is applied in a layer-wise manner over the injected layers $\ell \in \mathcal{L}$. 
At each layer, the input is the set of adapter updates $\{\Delta W_i^{(\ell)}\}_{i \in S}$, and the output is a same-cardinality set $\{\widetilde{\Delta W}_i^{(\ell)}\}_{i \in S}$ that can be directly used for joint inference.

As illustrated in Fig.~\ref{fig:wide}, the transformation is implemented as a structured pipeline consisting of three stages: 
(1) low-rank signature extraction, 
(2) a permutation-equivariant, conflict-aware set transformation, and 
(3) structured reconstruction into standard adapter updates.

The core transformation operates on the selected adapters as an unordered set, while preserving the correspondence between inputs and outputs. 

\subsection{Low-Rank Signature Space}
\label{sec:5.2}

Each adapter update is represented in a compact low-rank signature space (Fig.~\ref{fig:wide}B), retaining both dominant output directions and their magnitudes for subsequent transformation.

For adapter $i \in S$ at layer $\ell \in \mathcal{L}$, the update $\Delta W_i^{(\ell)} \in \mathbb{R}^{d_\ell \times k_\ell}$ is approximated by a rank-$K$ truncated singular value decomposition:
\begin{equation}
  \Delta W_i^{(\ell)}
  \approx
  U_i^{(\ell)} \Sigma_i^{(\ell)} V_i^{(\ell)\top}
  \label{eq:truncated_svd}
\end{equation}
where $U_i^{(\ell)} \in \mathbb{R}^{d_\ell \times K}$ captures the dominant output directions, $\Sigma_i^{(\ell)}$ encodes their magnitudes, and $V_i^{(\ell)} \in \mathbb{R}^{k_\ell \times K}$ represents the input-side basis. In our formulation, the transformation is applied only to the output-side factor $U_i^{(\ell)} \Sigma_i^{(\ell)}$, while $V_i^{(\ell)}$ is kept fixed. This choice preserves the input-side selectivity already encoded by each adapter and avoids simultaneously modifying both the read-in and write-out structure of the update. In the injected layers considered here, the incoming hidden states are already strongly shared across jointly loaded adapters, so incompatibility is dominated by how adapters write into common output directions. Fixing $V_i^{(\ell)}$ therefore makes the rewriting more constrained and reduces the risk of over-deconflicting adapters that are already separated by their input-side responses.

The output-side behavior is summarized by $B_i^{(\ell)} = U_i^{(\ell)} \Sigma_i^{(\ell)}$, which retains both the dominant output directions and their associated magnitudes. It is then vectorized as $s_i^{(\ell)} = \mathrm{flat}(B_i^{(\ell)}) \in \mathbb{R}^{d_\ell K}$, providing a consistent representation across adapters for set-level processing. At layer $\ell$, the selected adapters are represented as a set of signatures $\mathcal{X}^{(\ell)} = \{s_i^{(\ell)}\}_{i \in S}$.

\subsection{Subspace-Deconflicting Set Transformation}

Given $\mathcal{X}^{(\ell)}$, the core of SDO is a permutation-equivariant set operator $T_{\psi}^{(\ell)}$ that jointly updates all adapter signatures at layer $\ell$ (Fig.~\ref{fig:wide}C). This operator rewrites each element based on its own state, the surrounding set, and pairwise subspace conflicts, while preserving the input set's unordered structure.

Each signature is first mapped to an embedded representation $z_i^{(\ell)} = \phi_\psi^{(\ell)}(s_i^{(\ell)})$. To explicitly model geometric incompatibility, we introduce a pairwise conflict prior:
\begin{equation}
  \rho_{ij}^{(\ell)}
  =
  \left\|
  U_i^{(\ell)\top} U_j^{(\ell)}
  \right\|_F^2
  \label{eq:conflict_score}
\end{equation}
where $U_i^{(\ell)}$ and $U_j^{(\ell)}$ are the dominant output-side bases obtained from the truncated SVD in Eq.~\ref{eq:truncated_svd}. This prior captures geometric overlap between dominant output subspaces and serves as a lightweight pairwise cue for set interaction. Magnitude-aware conflict is handled later by the structure-level objectives.

Based on the embedded representations, query, key, and value vectors are constructed as $q_i^{(\ell)} = W_q^{(\ell)} z_i^{(\ell)}$, $k_i^{(\ell)} = W_k^{(\ell)} z_i^{(\ell)}$, and $v_i^{(\ell)} = W_v^{(\ell)} z_i^{(\ell)}$. Learned affinity and the geometric conflict prior are then combined to define interaction weights:
\begin{equation}
  a_{ij}^{(\ell)}
  =
  \frac{
    \exp\!\left(
      \frac{q_i^{(\ell)\top} k_j^{(\ell)}}{\sqrt{d_z}}
      + \beta \rho_{ij}^{(\ell)}
    \right)
  }{
    \sum\limits_{t\in S,\ t\neq i}
    \exp\!\left(
      \frac{q_i^{(\ell)\top} k_t^{(\ell)}}{\sqrt{d_z}}
      + \beta \rho_{it}^{(\ell)}
    \right)
  }
  \quad j \ne i
  \label{eq:attention_weight}
\end{equation}
where $d_z$ is the embedding dimension, $\beta \ge 0$ controls the strength of the conflict prior, and $a_{ii}^{(\ell)} = 0$. For fixed learned affinity, larger subspace overlap increases the interaction weight, encouraging the operator to attend more strongly to potentially conflicting adapters during context-aware correction.

The interaction weights define two context terms. The first is a local message, $m_i^{(\ell)} = \sum_{j\in S,\ j\neq i} a_{ij}^{(\ell)} v_j^{(\ell)}$, and the second is a global context, $g^{(\ell)} = \frac{1}{|S|}\sum_{t\in S} z_t^{(\ell)}$. The updated signature is then obtained through a residual correction:
\begin{equation}
  \delta_i^{(\ell)} = f_{\psi}^{(\ell)}\!\left([\,s_i^{(\ell)}; m_i^{(\ell)}; g^{(\ell)}\,]\right),
  \quad
  \hat{s}_i^{(\ell)} = s_i^{(\ell)} + \delta_i^{(\ell)}
  \label{eq:updated_signature}
\end{equation}
where $f_{\psi}^{(\ell)}$ is an MLP mapping to $\mathbb{R}^{d_\ell K}$. This residual design keeps the original signature as the reference state and applies only a context-dependent correction, helping preserve adapter-specific behavior while enabling conflict-aware adjustment under multi-adapter interaction.

\subsection{Reconstruction and Joint Inference}

As illustrated in Fig.~\ref{fig:wide}D, each updated signature is reshaped to $\hat{B}_i^{(\ell)} = \mathrm{unflat}(\hat{s}_i^{(\ell)}) \in \mathbb{R}^{d_\ell \times K}$, followed by a right polar factorization:
\begin{equation}
  \hat{B}_i^{(\ell)} = Q_i^{(\ell)} H_i^{(\ell)},
  \quad
  Q_i^{(\ell)\top}Q_i^{(\ell)} = I_K,
  \quad
  H_i^{(\ell)} \succeq 0
  \label{eq:polar}
\end{equation}
where $Q_i^{(\ell)} \in \mathbb{R}^{d_\ell \times K}$ has orthonormal columns and defines the transformed output subspace, and $H_i^{(\ell)} \in \mathbb{R}^{K \times K}$ is a symmetric positive semi-definite matrix that captures its magnitude structure.

Using the original right basis $V_i^{(\ell)}$ from Eq.~\ref{eq:truncated_svd}, the transformed update is reconstructed as
$\widetilde{\Delta W}_i^{(\ell)} = Q_i^{(\ell)} H_i^{(\ell)} V_i^{(\ell)\top}$.
Since $H_i^{(\ell)} \succeq 0$, the reconstructed update remains a valid low-rank update and can be directly used in standard adapter-based inference pipelines.

SDO returns a set of adapters with the same cardinality rather than a merged module. Joint inference follows the standard multi-adapter loading rule, with an effective update
$\Delta W_{\mathrm{eff}}^{(\ell)}=\sum_{i\in S}\alpha_i \widetilde{\Delta W}_i^{(\ell)}$,
where $\alpha_i$ is the runtime loading scale.

\begin{table*}[htbp]
  \centering
  \small
  \setlength{\tabcolsep}{3.5pt}
  \renewcommand{\arraystretch}{1.08}
  \begin{tabular}{ll|cccc|cccc|cccc|cccc}
  \hline
  \multirow{2}{*}{Setting} & \multirow{2}{*}{Method}
  & \multicolumn{4}{c|}{2 LoRAs}
  & \multicolumn{4}{c|}{3 LoRAs}
  & \multicolumn{4}{c|}{4 LoRAs}
  & \multicolumn{4}{c}{5 LoRAs} \\
  \cline{3-18}
  &
  & ID$\uparrow$ & IR$\uparrow$ & FCS$\uparrow$ & CL$\uparrow$
  & ID$\uparrow$ & IR$\uparrow$ & FCS$\uparrow$ & CL$\uparrow$
  & ID$\uparrow$ & IR$\uparrow$ & FCS$\uparrow$ & CL$\uparrow$
  & ID$\uparrow$ & IR$\uparrow$ & FCS$\uparrow$ & CL$\uparrow$ \\
  \hline

  \multirow{5}{*}{SM-FLUX}
  & Naive Merge & 0.6107 & 1.00 & 1.00 & 0.2669 & 0.5425 & 1.00 & 1.00 & 0.1779 & 0.5637 & 1.00 & 1.00 & 0.2382 & 0.4805
  & 0.60 & 0.80 & 0.2250 \\
  & CharCom     & 0.6974 & 1.00 & 1.00 & \textbf{0.2966} & 0.5462 & 1.00 & 1.00 & 0.1929 & 0.5575 & 1.00 & 1.00 &
  \textbf{0.2565} & 0.5006 & 0.80 & 1.00 & \textbf{0.2924} \\
  & FreeFuse    & \textbf{0.7847} & 1.00 & 1.00 & 0.2395 & 0.5991 & 1.00 & 1.00 & 0.1962 & 0.5564 & 0.75 & 1.00 & 0.2298
  & 0.5038 & 0.60 & 0.80 & 0.2274 \\
  & LoRA-Composer   & 0.7421 & 1.00 & 1.00 & 0.2487 & 0.5874 & 1.00 & 1.00 & 0.2056 & 0.5338 & 0.75 & 1.00 & 0.2215 & 0.4916
  & 0.60 & 0.80 & 0.2307 \\
  & SDO         & 0.7634 & 1.00 & 1.00 & 0.2523 & \textbf{0.6682} & 1.00 & 1.00 & \textbf{0.2091}
               & \textbf{0.6680} & 1.00 & 1.00 & 0.2447
               & \textbf{0.6565} & \textbf{1.00} & \textbf{1.00} & 0.2537 \\
  \hline

  \multirow{5}{*}{FL-FLUX}
  & Naive Merge & 0.7318 & 1.00 & 1.00 & 0.3010 & 0.7426 & 1.00 & 1.00 & 0.3324 & 0.6915 & 1.00 & 1.00 & 0.2746 & 0.6428
  & 1.00 & 1.00 & 0.2481 \\
  & CharCom     & \textbf{0.7603} & 1.00 & 1.00 & 0.2522 & 0.7578 & 1.00 & 1.00 & 0.2791 & \textbf{0.7442} & 1.00 & 1.00 & 0.2242
  & 0.6997 & 1.00 & 1.00 & 0.2500 \\
  & FreeFuse    & 0.7489 & 1.00 & 1.00 & 0.2894 & 0.7531 & 1.00 & 1.00 & 0.3187 & 0.7124 & 1.00 & 1.00 & 0.2619 & 0.6645
  & 1.00 & 1.00 & 0.2396 \\
  & LoRA-Composer   & 0.7396 & 1.00 & 1.00 & 0.2815 & 0.7468 & 1.00 & 1.00 & 0.3093 & 0.7041 & 1.00 & 1.00 & 0.2498 & 0.6517
  & 1.00 & 1.00 & 0.2324 \\
  & SDO         & 0.7454 & 1.00 & 1.00 & \textbf{0.3385}
             & \textbf{0.7504} & 1.00 & 1.00 & \textbf{0.3542}
             & 0.7339 & 1.00 & 1.00 & \textbf{0.2867}
             & \textbf{0.7101} & 1.00 & 1.00 & \textbf{0.2633} \\
  \hline

  \multirow{5}{*}{SM-SDXL}
  & Naive Merge & 0.5039 & 1.00 & 1.00 & 0.2593 & 0.4993 & 1.00 & 1.00 & 0.1745 & 0.4833 & 0.75 & 1.00 & 0.2917 & 0.3722
  & 0.60 & 0.80 & 0.2233 \\
  & CharCom     & 0.5573 & 1.00 & 1.00 & 0.3349 & 0.5823 & 1.00 & 1.00 & 0.2001 & 0.4722 & 1.00 & 1.00 & 0.2212 & 0.4413
  & 0.80 & 1.00 & 0.2658 \\
  & FreeFuse    & 0.5314 & 1.00 & 1.00 & \textbf{0.3401} & 0.5612 & 1.00 & 1.00 & 0.2153 & 0.4837 & 1.00 & 1.00 & 0.2820
  & 0.4622 & 0.80 & 1.00 & 0.2468 \\
  & LoRA-Composer   & 0.5491 & 1.00 & 1.00 & 0.3188 & 0.5734 & 1.00 & 1.00 & 0.2065 & 0.4618 & 0.75 & 1.00 & 0.2367 & 0.4386
  & 0.60 & 0.80 & 0.2299 \\
  & SDO         & \textbf{0.5668} & 1.00 & 1.00 & 0.3204
             & \textbf{0.5797} & 1.00 & 1.00 & \textbf{0.2249}
             & \textbf{0.5314} & 1.00 & 1.00 & \textbf{0.2972}
             & \textbf{0.5161} & \textbf{1.00} & \textbf{1.00} & \textbf{0.2715} \\
  \hline
  \end{tabular}
\caption{Main quantitative results under three evaluation settings: in-domain (Shakoomaku on FLUX-dev), cross-topic (Flintstones on FLUX-dev), and cross-backbone (Shakoomaku on SDXL). `SM' and `FL' denote the Shakoomaku and Flintstones LoRA pools, respectively. All methods are evaluated under identical adapter subsets, prompts, random seeds, and inference configurations. Performance differences become more noticeable in higher-cardinality settings (4--5 LoRAs), where multi-adapter composition is more challenging. Higher is better for all metrics; ties in saturated IR/FCS columns are not additionally bolded.}
  \label{tab:main_results_all}
  \end{table*}

\subsection{Training Objective and Optimization}

SDO is trained offline and then deployed as a plug-and-play operator. At test time, a trained SDO rewrites each selected adapter set in a single forward pass without per-composition optimization. The overall training objective combines task-level identity supervision with two structure-level regularizers:
\begin{equation}
  \mathcal{L}
  =
  \lambda_{id}\mathcal{L}_{id}
  +
  \lambda_{conf}\mathcal{L}_{conf}
  +
  \lambda_{pres}\mathcal{L}_{pres}
  \label{eq:total_loss}
\end{equation}
where $\lambda_{id}, \lambda_{conf}, \lambda_{pres} \ge 0$ balance identity supervision, conflict reduction, and geometry preservation. Whether $\mathcal{L}_{id}$ contributes gradients during optimization depends on the identity backend, as specified below.

The identity term encourages the transformed adapter set to preserve the intended subjects after image generation. For each adapter $i$, $N_i$ reference images $\{r_{i,t}\}_{t=1}^{N_i}$ are encoded by a frozen identity encoder $E_{id}$ to form a prototype
$\bar{p}_i = \frac{1}{N_i}\sum_{t=1}^{N_i}
\frac{E_{id}(r_{i,t})}{\|E_{id}(r_{i,t})\|_2}$ and
$p_i = \frac{\bar{p}_i}{\|\bar{p}_i\|_2}$. Given a training prompt, the frozen diffusion model renders an image $y$ using the effective multi-adapter update. A subject-instance backend extracts candidate subject instances $\mathcal{D}(y)=\{d_j\}_{j=1}^{n_y}$, each encoded as
$u_j = \frac{E_{id}(d_j)}{\|E_{id}(d_j)\|_2}$.

The matching cost is $c_{ij} = 1 - p_i^\top u_j$, and a minimum-cost partial one-to-one assignment $\pi^\star \subseteq S \times \{1,\dots,n_y\}$ is computed between expected identities and detected subject instances, yielding:
\begin{equation}
  \mathcal{L}_{id}
  =
  \frac{1}{\max(1,|\pi^\star|)}
  \sum_{(i,j)\in\pi^\star}
  \left(1 - p_i^\top u_j\right)
  +
  \lambda_{miss}
  \frac{\bigl|\,|\mathcal{D}(y)| - |S|\,\bigr|}{|S|}
  \label{eq:id_loss}
\end{equation}
where the first term penalizes identity mismatch and the second penalizes missing or extra detected subject instances, weighted by $\lambda_{miss} \ge 0$. In this work, the backend is instantiated with a face-oriented detector/encoder because the evaluated benchmarks are identity-centric human or character datasets. The formulation itself, however, only assumes subject-level instance localization and can be paired with alternatives such as cross-attention-based localization or open-vocabulary detectors for non-face subjects.

Two regularizers act directly in parameter space. While Eq.~\eqref{eq:conflict_score} provides a lightweight geometric prior for set interaction, the energy-aware suppression of harmful overlap is enforced after transformation through the following conflict loss:
\begin{equation}
  \mathcal{L}_{conf}
  =
  \frac{1}{|\mathcal{L}|\binom{|S|}{2}}
  \sum_{\ell\in\mathcal{L}}
  \sum_{\substack{i,j\in S\\ i<j}}
  \left\|
  H_i^{(\ell)\frac{1}{2}}
  Q_i^{(\ell)\top}
  Q_j^{(\ell)}
  H_j^{(\ell)\frac{1}{2}}
  \right\|_F^2
  \label{eq:conf_loss}
\end{equation}
This loss averages pairwise overlap energy across layers and adapter pairs in the selected set. To preserve the original geometry, the preservation loss is defined as:
\begin{equation}
  \begin{aligned}
    \mathcal{L}_{pres}
    &=
    \frac{1}{|\mathcal{L}|\,|S|}
    \sum_{\ell\in\mathcal{L}}
    \sum_{i\in S}
    \Bigl(
      \left\|
      Q_i^{(\ell)}Q_i^{(\ell)\top}
      -
      U_i^{(\ell)}U_i^{(\ell)\top}
      \right\|_F^2 \\
    &\qquad\qquad\qquad\qquad
      + \gamma
      \left\|
      H_i^{(\ell)}
      -
      \Sigma_i^{(\ell)}
      \right\|_F^2
    \Bigr)
  \end{aligned}
  \label{eq:pres_loss}
\end{equation}
where the first term preserves the output subspace and the second keeps the magnitude structure close to the original singular values, ensuring that the transformed updates remain close to the originals in both direction and magnitude. The coefficient $\gamma$ controls the balance between the two terms.

The identity backend may be differentiable or non-differentiable. When it is differentiable, gradients from $\mathcal{L}_{id}$ propagate through the rendering process to the SDO operator. Under a non-differentiable external subject-instance backend, $\mathcal{L}_{id}$ is still evaluated but used only for monitoring and checkpoint selection, while gradient-based parameter updates are driven by $\mathcal{L}_{conf}$ and $\mathcal{L}_{pres}$.

\begin{table}[!htbp]
\centering
\small
\setlength{\tabcolsep}{3pt}
\renewcommand{\arraystretch}{1.05}
\begin{tabular}{lcccccccc}
\hline
\multirow{2}{*}{Variant}
& \multicolumn{4}{c}{4 LoRAs}
& \multicolumn{4}{c}{5 LoRAs} \\
\cmidrule(lr){2-5}\cmidrule(lr){6-9}
& ID$\uparrow$ & IR$\uparrow$ & FCS$\uparrow$ & CL$\uparrow$
& ID$\uparrow$ & IR$\uparrow$ & FCS$\uparrow$ & CL$\uparrow$ \\
\hline

Full SDO
& 0.6680 & 1.00 & 1.00 & 0.2447
& 0.6565 & 1.00 & 1.00 & 0.2537 \\

w/o CP
& 0.6512 & 0.75 & 1.00 & 0.2398
& 0.5676 & 0.80 & 0.80 & 0.2459 \\

w/o GC
& 0.6404 & 1.00 & 1.00 & 0.2429
& 0.6098 & 0.80 & 1.00 & 0.2488 \\

w/o SR
& 0.6241 & 0.75 & 0.75 & 0.2416
& 0.5913 & 0.80 & 0.60 & 0.2474 \\

\hline
\end{tabular}
\caption{Ablation study of SDO under 4- and 5-LoRA composition. Removing the conflict prior (CP), global context (GC), or structured reconstruction (SR) consistently degrades performance, especially in the more challenging 5-LoRA setting. The largest drop occurs when CP is removed, highlighting the importance of explicitly modeling subspace conflict.}
\label{tab:ablation_components}
\end{table}

During training, the base diffusion model and all source adapters remain frozen, and only the SDO parameters $\psi$ are optimized. Training proceeds in two stages. The first optimizes $\mathcal{L}_{conf} + \mathcal{L}_{pres}$ to initialize a geometry-preserving operator. The second continues from this initialization under Eq.~\eqref{eq:total_loss}, with the effective gradient path determined by the identity backend. At each step, the SDO operator is applied to a selected adapter set, the rewritten parameters are used to evaluate the structure-level losses, and a rendered image is additionally used to evaluate $\mathcal{L}_{id}$ when applicable. Once optimized, the learned SDO parameters are reused across deployment-time adapter subsets without test-time optimization.

\begin{figure*}[htbp]
    \centering
    \includegraphics[width=0.8\textwidth, keepaspectratio]{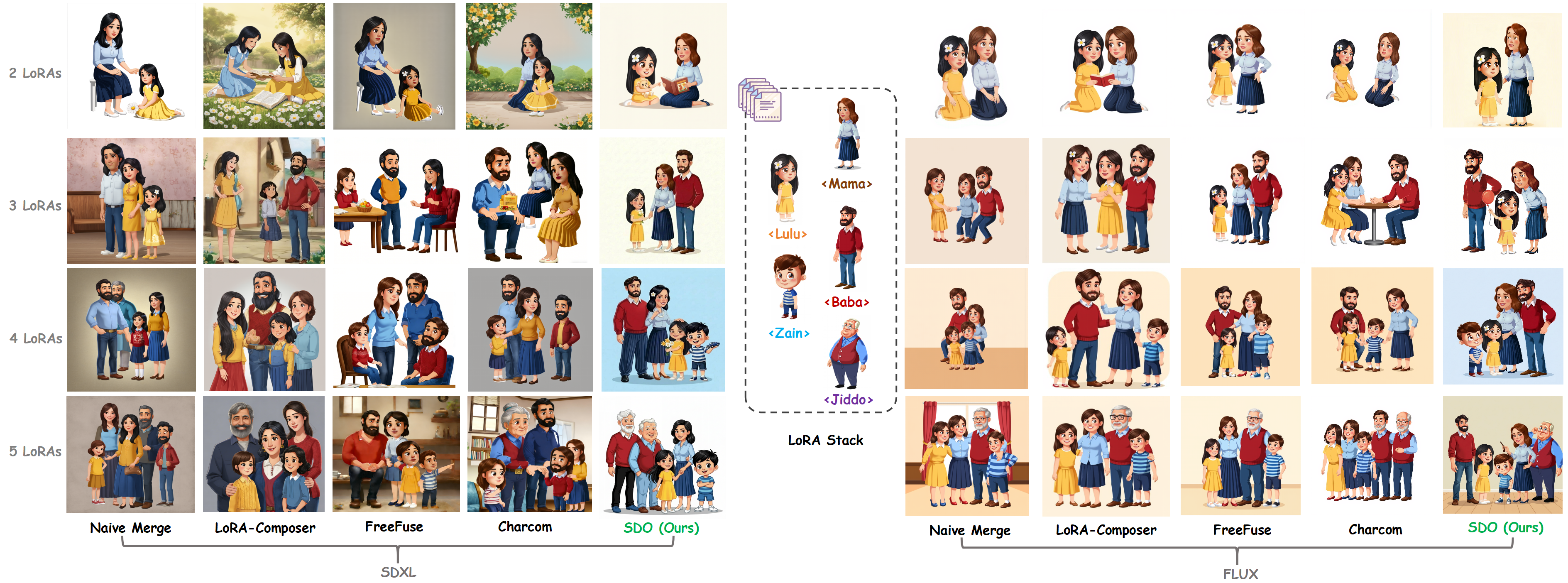}
    \caption{Qualitative comparison under progressively increasing scene complexity. As more adapters are jointly loaded, baseline methods tend to show stronger identity blending, missing characters, and cross-character attribute leakage. In contrast, SDO better maintains subject identity and separation, yielding more coherent multi-character compositions, with differences becoming more noticeable in the 4- and 5-LoRA settings.}
    \label{fig:rotated}
\end{figure*}

\section{Experiments}

\subsection{Experimental Setup}

We use two identity-specific LoRA pools, \textbf{Shakoomaku (SM)} and \textbf{Flintstones (FL)}, each containing five independently trained identity adapters. We use \textbf{FLUX.1-dev} and \textbf{SDXL-base-1.0} as the two diffusion backbones in our experiments, and refer to them as \textbf{FLUX} and \textbf{SDXL} in the remainder of the paper. SDO is trained only on FLUX. We evaluate three settings: \textbf{in-domain} (SM-FLUX), \textbf{cross-topic} (FL-FLUX), and \textbf{cross-backbone} (SM-SDXL), where the last setting tests direct transfer without retraining.

We study controlled few-adapter composition: for each setting, we evaluate 2-, 3-, 4-, and 5-LoRA composition under identical adapter subsets, prompts, random seeds, and inference configurations across methods. We compare SDO against four representative baselines: \textbf{Naive Merge}, \textbf{CharCom}~\cite{wang2025charcom}, \textbf{FreeFuse}~\cite{liu2025freefuse}, and \textbf{LoRA-Composer}~\cite{yang2025loracomposer}. Performance is assessed from the perspectives of identity preservation and overall generation quality using four metrics: identity similarity (ID), identity recall (IR), face count score (FCS), and CLIP similarity (CL). ID measures similarity between generated and reference identities, IR measures how many intended identities are successfully preserved, FCS measures whether the correct number of faces is rendered, and CL measures image-text alignment.

\subsection{Main Results on Multi-LoRA Generalization}

Table~\ref{tab:main_results_all} reports the quantitative results. In the easier 2- and 3-LoRA regimes, SDO remains competitive with strong baselines, with most observable differences concentrated in ID, while IR and FCS are often saturated. This indicates that the proposed deconflicting operator does not compromise generation quality when inter-adapter conflict is still limited. As composition becomes more challenging, the advantage of SDO becomes more evident. On the in-domain SM-FLUX and cross-backbone SM-SDXL settings, SDO achieves stronger identity preservation and more stable composition in the 4- and 5-LoRA regimes, providing preliminary evidence that the learned deconfliction strategy transfers across backbones.

In the FL-FLUX setting, the margin is more modest, but SDO remains competitive and continues to show the largest gains in the most difficult 4- and 5-LoRA cases. This behavior is consistent with weaker identity specificity and greater intra-identity variation in this pool, leading to less pronounced improvements under the current metrics. Overall, SDO is most beneficial in high-complexity composition regimes, while remaining comparable to strong baselines in easier cases and without sacrificing text–image alignment.

\subsection{Ablation Study}

We perform ablations on key design components of SDO on SM-FLUX, focusing on the more challenging 4- and 5-LoRA regimes where multi-adapter interference is most pronounced. Specifically, we consider removing the conflict prior in the operator (\textbf{w/o CP}), removing the global set context (\textbf{w/o GC}), and replacing the structured reconstruction with a direct reconstruction scheme (\textbf{w/o SR}).

Table~\ref{tab:ablation_components} shows that the full SDO achieves the best overall performance, with larger gaps emerging in the 5-LoRA setting. Removing \textbf{CP} results in the clearest drop in identity-related metrics, indicating that explicitly modeling pairwise subspace conflict is critical for effective deconfliction within the operator. Removing \textbf{GC} also degrades performance, suggesting that pairwise interactions alone are insufficient without set-level context. Replacing \textbf{SR} primarily affects compositional stability, with a pronounced decline in face-count consistency, highlighting the role of structured reconstruction in maintaining stable multi-character generation. Across all variants, CL remains largely unchanged, indicating that these components primarily impact identity preservation and multi-character consistency rather than text–image alignment.

\subsection{Qualitative Evaluation, Human Study, and Subspace Analysis}

\paragraph{Qualitative Results.}

The qualitative comparisons in Fig.~\ref{fig:teaser} and Fig.~\ref{fig:rotated} align with the quantitative trends. As the number of composed adapters increases, baseline methods exhibit more severe identity inconsistency, including identity blending, missing characters, and cross-character attribute leakage. In contrast, SDO maintains clearer identity separation and more coherent scene composition, with differences becoming more pronounced in the crowded 4- and 5-LoRA cases. On the FL-FLUX setting, the visual gap is less pronounced, consistent with weaker identity-specific cues and higher intra-identity variation. These observations are consistent with the view that reducing subspace conflict leads to more stable and identity-consistent multi-adapter composition.

\begin{figure}[t]
    \centering
    \includegraphics[width=\linewidth]{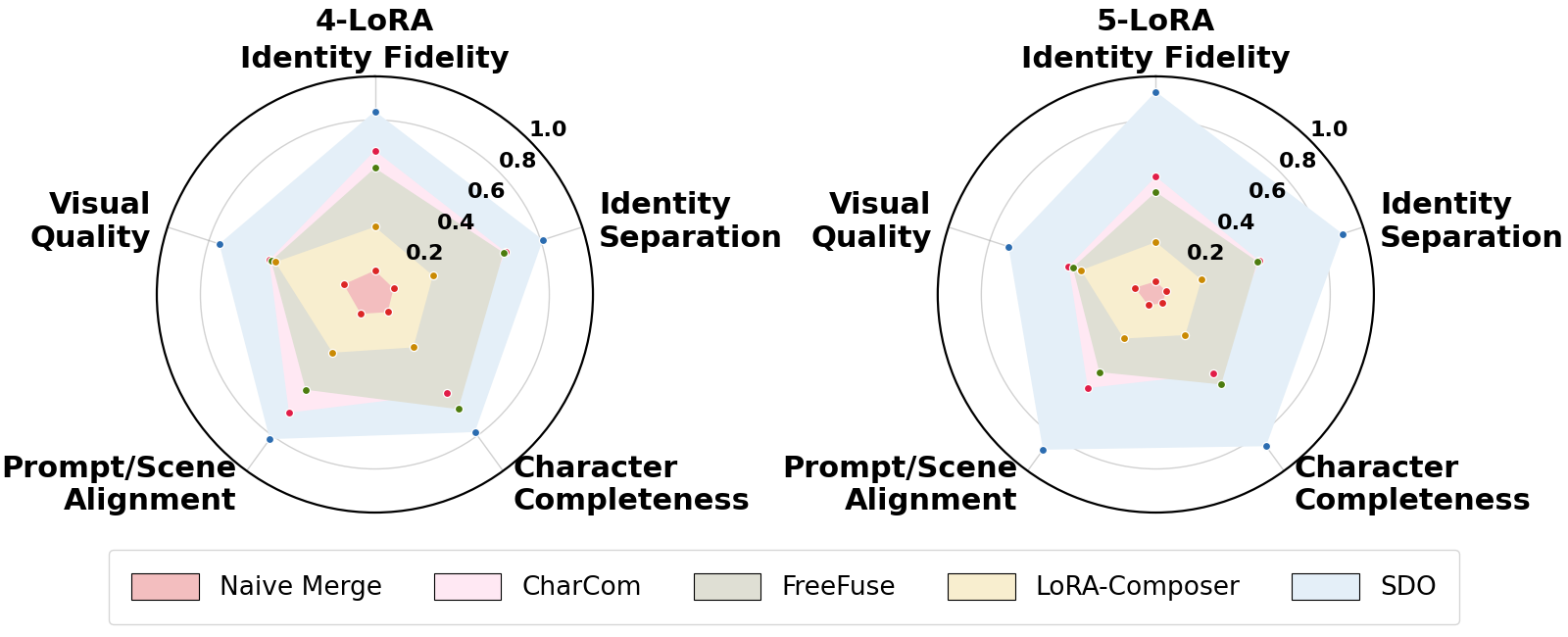}
    \caption{Human evaluation results visualized as radar plots for 4- and 5-LoRA settings. Each axis corresponds to a different criterion: identity fidelity, identity separation, character completeness, prompt/scene alignment, and visual quality. Different colors represent different methods. SDO achieves the strongest overall performance, especially in the 5-LoRA setting.}
    \label{fig:human_eval}
\end{figure}

\paragraph{Human Evaluation.}
We conduct human evaluation on the more challenging 4- and 5-LoRA settings using pairwise preference comparisons. For each prompt and criterion, annotators are presented with two images and asked to select the preferred one. Preferences are aggregated into per-method win rates across five criteria: identity fidelity (IF), identity separation (IS), character completeness (CC), prompt/scene alignment (PSA), and visual quality (VQ). As shown in Fig. 5, SDO consistently achieves the strongest overall performance. The advantage is already clear in the 4-LoRA setting and becomes more pronounced in the 5-LoRA case, indicating improved identity consistency and perceptual quality as composition complexity increases.

\paragraph{Subspace Analysis.}

To directly examine whether SDO reduces parameter-space conflict, we visualize adapter signatures before and after deconfliction under FLUX and SDXL. For each adapter, layer-wise behavior signatures are concatenated across shared layers to form a single adapter-level representation, so that each adapter corresponds to one point in the visualization. Within each backbone, the union of original and transformed representations is projected into a shared 2D PCA space, while pairwise overlap is measured in the original high-dimensional signature space.

\begin{figure}[t]
    \centering
    \includegraphics[width=\linewidth]{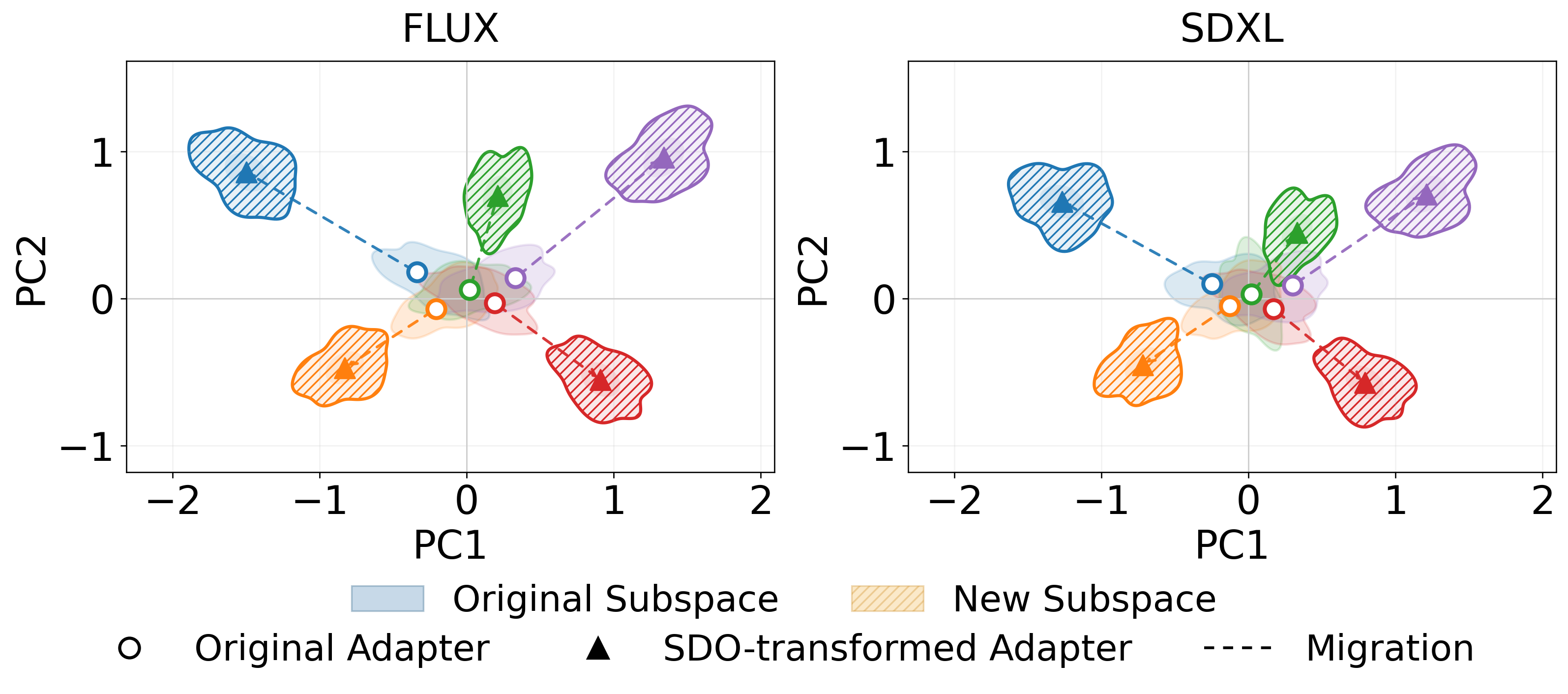}
    \caption{PCA visualization of adapter signatures before and after SDO under FLUX and SDXL. Each point represents one adapter, formed by concatenating layer-wise behavior signatures across shared layers. Circles denote original adapters and triangles denote the SDO-transformed version.}
    \label{fig:lora_projection_backbones}
\end{figure}

As shown in Fig.~\ref{fig:lora_projection_backbones}, SDO-transformed adapters exhibit clearer separation on both backbones, and this visual pattern is consistent with the reduced average pairwise overlap measured in the original signature space. Although the figure is only a 2D projection, it provides an intuitive geometric view suggesting that SDO reduces conflict among the selected adapters in parameter space. This geometric trend is also consistent with the larger gains observed in the more challenging 4- and 5-adapter settings.

\section{Conclusion}

We formulate multi-adapter interference as a subspace-conflict problem in parameter space and introduce SDO, a learnable permutation-equivariant operator that performs conflict-aware rewriting before composition. By operating in a structured low-rank representation and reconstructing the updates into deployable adapter parameters, SDO provides an effective approach to improving identity-consistent multi-adapter generation. The improvements are most pronounced in challenging 4- and 5-adapter regimes, indicating that explicit subspace-level reasoning is particularly beneficial under high compositional complexity. At the same time, the current study is limited to controlled few-adapter composition with two five-identity LoRA pools, and extending this framework to larger-scale settings and more diverse adapter collections remains an important direction for future work.
\bibliographystyle{ACM-Reference-Format}
\balance
\bibliography{sample-base}


\end{document}